\documentclass[acmtog,nonacm]{acmart}

\AtBeginDocument{%
  \providecommand\BibTeX{{%
    \normalfont B\kern-0.5em{\scshape i\kern-0.25em b}\kern-0.8em\TeX}}}

\usepackage{epsfig}
\usepackage{graphicx}
\usepackage{amsmath}
\usepackage{algorithm}
\usepackage{algorithmic}
\usepackage{mathbbol}
\usepackage{multirow}

\newcommand{\eg}[1]{e.g.,}
\newcommand{\ie}[1]{i.e.,}
\newcommand{\etal}[1]{et al.}

\usepackage{color}
\newcommand{\yeA}[1]{#1}

\setcopyright{none}
\renewcommand\footnotetextcopyrightpermission[1]{}
\usepackage{tabularx} 

\begin{document}

\title{
Masked-Attention Diffusion Guidance for Spatially Controlling Text-to-Image Generation
}

\author{Yuki Endo}
\affiliation{%
  \institution{University of Tsukuba}
  \city{Ibaraki}
  \country{Japan}
}
\email{endo@cs.tsukuba.ac.jp}
\orcid{0000-0001-5132-3350}


\begin{abstract}
Text-to-image synthesis has achieved high-quality results with recent advances in diffusion models. However, text input alone has high spatial ambiguity and limited user controllability. Most existing methods allow spatial control through additional visual guidance (e.g., sketches and semantic masks) but require additional training with annotated images. In this paper, we propose a method for spatially controlling text-to-image generation without further training of diffusion models. Our method is based on the insight that the cross-attention maps reflect the positional relationship between words and pixels. Our aim is to control the attention maps according to given semantic masks and text prompts. To this end, we first explore a simple approach of directly swapping the cross-attention maps with constant maps computed from the semantic regions. 
\yeA{
Some prior works also allow training-free spatial control of text-to-image diffusion models by directly manipulating cross-attention maps. However, these approaches still suffer from misalignment to given masks because manipulated attention maps are far from actual ones learned by diffusion models. To address this issue, 
}
we propose masked-attention guidance, which can generate images more faithful to semantic masks \yeA{via indirect control of} attention to each word and pixel by manipulating noise images fed to diffusion models. 
\yeA{
Masked-attention guidance can be easily integrated into pre-trained off-the-shelf diffusion models (e.g., Stable Diffusion) and applied to the tasks of text-guided image editing.
}
Experiments show that our method enables more accurate spatial control than baselines qualitatively and quantitatively. The code is available at \href{https://github.com/endo-yuki-t/MAG}{\color{magenta}{https://github.com/endo-yuki-t/MAG}}.
\end{abstract}

\keywords{Diffusion Model, Text-to-Image Synthesis, Multimodal, Classifier Guidance}

\begin{teaserfigure}
  \includegraphics[width=\textwidth]{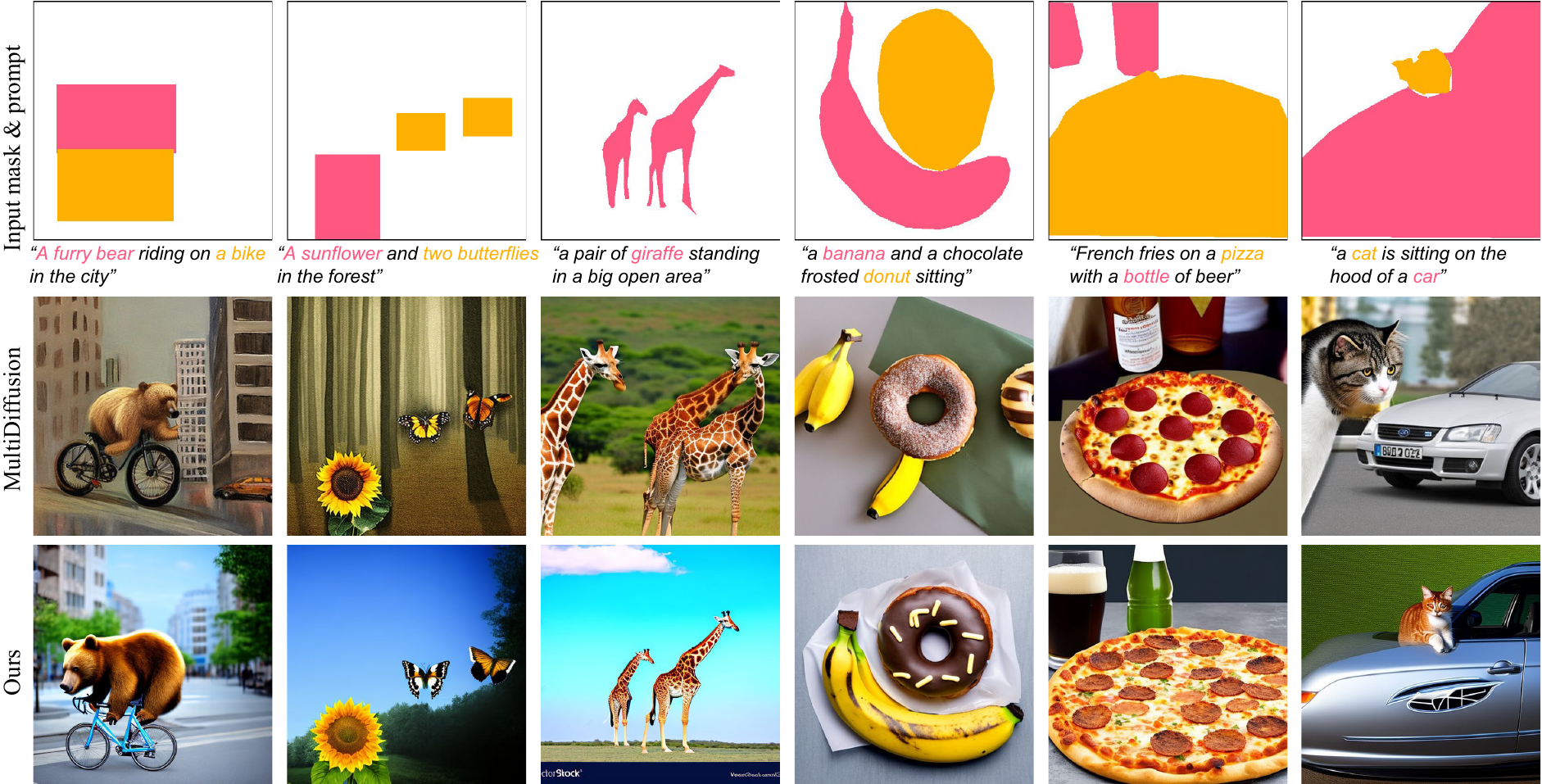}
  \caption{
  Spatially controlling text-to-image generation using visual guidance without additional training of Stable Diffusion~\cite{Rombach_2022_CVPR}. Our method can generate images more faithful to given semantic masks than the state-of-the-art method (MultiDiffusion)~\cite{bar2023multidiffusion}. 
  }
  \label{fig:teaser}
\end{teaserfigure}


\maketitle
\section{Introduction}
\label{sec:Introduction}
Text-to-image generation techniques have rapidly evolved thanks to the advances in image-generative models and language models. In particular, diffusion model-based methods, such as Imagen~\cite{DBLP:conf/nips/SahariaCSLWDGLA22} and Stable Diffusion~\cite{Rombach_2022_CVPR}, achieve state-of-the-art quality. Although text input is helpful for intuitive control, it has spatial ambiguity and does not allow sufficient user control. For example, from a text prompt, "\textit{A furry bear riding on a bike in the city}," we cannot identify spatial attributes like the shape, size, and position of the bear. Therefore, controlling text-to-image generation with multimodal inputs, including bounding boxes and semantic masks, is essential~\cite{zhan2021multimodal}. 

A straightforward way of multimodal control is to train diffusion models from scratch with annotated images in a supervised manner, but this requires much cost for annotation and training. ControlNet~\cite{DBLP:journals/corr/abs-2302-05543} adds layers to a trained diffusion model and trains only those layers, enabling control with sketches, poses, and semantic masks while reducing the training cost. Still, additional training takes hundreds of GPU hours and many annotated images. 
Bar-Tal et al. recently proposed a training-free approach, MultiDiffusion\yeA{~\shortcite{bar2023multidiffusion}}, which iteratively performs denoising for each semantic region using corresponding text input and merges the separately-denoised images. In the part of denoising steps, it adopts an approach called bootstrapping that gives fixed noise to regions other than the target semantic regions. This encourages semantic objects to be generated in the specified regions. 
However, during bootstrapping, images are generated independently for each semantic region without considering global context, sometimes resulting in copy-paste-like images. For example, the second column from the left in Figure~\ref{fig:teaser} and the most left column in Figure~\ref{fig:res} show that it generates foreground objects and backgrounds with different styles. 
Moreover, MultiDiffusion still struggles to generate images aligned to especially fine-grained masks (right four columns in Figure~\ref{fig:teaser}). 

In this paper, we tackle a problem of spatially controlling text-to-image generation without additional training. As shown in Figure~\ref{fig:teaser}, our diffusion model-based method generates images using visual guidance such as bounding boxes and semantic masks. To do so, we leverage the insight that the positional relationship between pixels and prompt words appears in cross-attention maps of diffusion models. We first explore a straightforward approach that directly manipulates the cross-attention maps according to given semantic masks. This is done at each step of a reverse diffusion process. \yeA{Recent works also adopt a similar idea. For example,} Directed Diffusion~\cite{ma2023directed} strengthens \yeA{and weakens} cross-attention maps according bounding boxes provided as visual guidance. To extend the input to arbitrary-shaped masks, we present \textit{masked-attention swapping} that replaces cross-attention maps with constant maps generated from semantic regions. 

However, directly manipulated cross-attention maps are far from actual ones estimated by general text-to-image diffusion models. As a result, this approach causes misalignment with visual guidance. To address this issue, we propose \textit{masked-attention guidance}, which manipulates noise maps fed to a diffusion model instead of directly manipulating attention maps. Masked-attention guidance simply encourages attention to increase or decrease according to semantic regions, and we leverage the ability of well-trained diffusion models to determine how much the network attends to which pixels and words. Masked-attention guidance is easy to implement and integrate into \yeA{pre-trained off-the-shelf diffusion models (e.g., Stable Diffusion~\cite{Rombach_2022_CVPR})}. \yeA{It can also be applied to the task of text-guided image editing.} Quantitative and qualitative comparisons with baseline methods show that our method is simple yet effective for spatial control of text-to-image generation. 

\section{Related Work}
\paragraph{Image-generative models.}
With the development of deep learning techniques, various generative models have been proposed to create high-quality images. Generative adversarial networks (GANs)~\cite{NIPS2014_5ca3e9b1} enable high-quality image generation via adversarial learning using a generator and discriminator. There have been various GAN-based methods, such as PGGAN~\cite{DBLP:conf/iclr/KarrasALL18}, BigGAN~\cite{DBLP:conf/iclr/BrockDS19}, and StyleGAN~\cite{DBLP:conf/cvpr/KarrasLA19,DBLP:conf/cvpr/KarrasLAHLA20,Karras2021,DBLP:conf/siggraph/SauerS022}, that can generate high-quality and diverse images. These models can create images not only for specific domains (e.g., human faces and animals) but also for large-scale and diverse datasets in recent years~\cite{DBLP:conf/siggraph/SauerS022}.

Meanwhile, diffusion models have also attracted much attention because of their remarkable image-generation capabilities. A representative baseline is denoising diffusion probabilistic model (DDPM)~\cite{DBLP:conf/nips/HoJA20}. DDPM is based on the forward and reverse diffusion processes, which gradually add tiny noise to data or recover original data from the noised data, step-by-step. In the reverse diffusion process, a U-Net-based model~\cite{DBLP:conf/miccai/RonnebergerFB15} estimates noise from noisy images at each step. The subsequent study presented ablated diffusion model (ADM)~\cite{DBLP:conf/nips/DhariwalN21}, which can generate higher-resolution images than DDPM and outperformed existing GAN-based methods. ADM adopts a two-stage upsampling approach~\cite{pmlr-v139-nichol21a} and introduces classifier guidance, which manipulates noised images according to classifier outputs to make output images close to a specific class. In general, diffusion models take more time for inference than GANs, but many acceleration techniques, such as denoising diffusion implicit model (DDIM)~\cite{DBLP:conf/iclr/SongME21}, have been studied. Also, diffusion models show state-of-the-art results in the text-to-image generation task, as described below. 

\paragraph{Text-guided image generation. }
CLIP~\cite{DBLP:conf/icml/RadfordKHRGASAM21} is a powerful vision and language model that greatly expands the range of applications of image-generative models. CLIP is trained using large-scale data on the Web to obtain image and text features in a joint embedding space. 
Text-to-image generation using CLIP features and StyleGANs has been widely explored in recent studies~\cite{tedigan21,styleclip21,clip2stylegan22,stylegan-nada22,maniclip22,hairclip22}. For example, StyleCLIP~\cite{styleclip21} shows an approach that manipulates latent codes of StyleGAN so as to increase the similarity between the CLIP features of StyleGAN images and input text. \yeA{While these} GAN-based approaches show promising results in specific domains like human faces, \yeA{
GigaGAN \cite{DBLP:conf/cvpr/KangZ0PSPP23} recently scaled up GANs to generate high-quality images in broader domains. 
}

Diffusion models have \yeA{also} enabled high-quality text-to-image generation in general domains. GLIDE~\cite{DBLP:conf/icml/NicholDRSMMSC22} introduces CLIP guidance and classifier-free guidance~\cite{DBLP:journals/corr/abs-2207-12598} into diffusion models for text-to-image generation. Furthermore, unCLIP (DALL-E2)~\cite{DBLP:journals/corr/abs-2204-06125} uses CLIP features as an additional input of diffusion models. Very powerful text-to-image methods, such as Imagen~\cite{DBLP:conf/nips/SahariaCSLWDGLA22} and Latent Diffusion~\cite{Rombach_2022_CVPR}, were also proposed. In addition, there have been various applications for editing real images using text input~\cite{DBLP:journals/corr/abs-2208-01626,DBLP:journals/corr/abs-2210-11427,brooks2022instructpix2pix,DBLP:journals/corr/abs-2302-03027,DBLP:conf/cvpr/LugmayrDRYTG22,Avrahami_2022_CVPR,DBLP:conf/cvpr/KimKY22a}. 
\yeA{
Another hot topic is personalized text-to-image generation~\cite{DBLP:conf/cvpr/RuizLJPRA23,DBLP:conf/cvpr/KumariZ0SZ23}, which learns specific concepts by finetuning some parameters of pre-trained diffusion models using several images. 
}
However, because text input alone has spatial ambiguity, we aim to improve the controllability of text-to-image generation using additional visual guidance as input. 

\paragraph{Spatial control of image-generative models. }
Pixel2Style2pixel (pSp)~\cite{DBLP:conf/cvpr/RichardsonAPNAS21} is a method that can control GANs in an image-to-image translation framework. This is achieved by training an encoder that converts visual guidance (e.g., sketches and semantic masks) into latent codes, but paired data are required for training the encoder. There have also been many efforts to manipulate latent codes for spatial editing~\cite{DBLP:conf/cvpr/ShenGTZ20,DBLP:journals/tog/AbdalZMW21,gansteerability,DBLP:conf/iclr/SpingarnBM21,DBLP:conf/icml/VoynovB20,DBLP:journals/corr/abs-2004-02546,DBLP:journals/corr/abs-2007-06600,DBLP:conf/iccv/HeKS21,zhu2021lowrankgan,DBLP:journals/cgf/Endo22,pan2023draggan}, such as object movement, rotation, and zooming. However, these methods have not been applied to text-to-image generation and only handle specific domains. 

As a diffusion model-based approach, SDEdit~\cite{DBLP:conf/iclr/MengHSSWZE22} can generate realistic images from stroke paintings via an img2img framework that adds noise to the input and then subsequently performs denoising. However, this method can add details to colorized paintings but cannot be applied to semantic masks. ControlNet~\cite{DBLP:journals/corr/abs-2302-05543} combines a pre-trained diffusion model with additional modules consisting of zero convolution and trainable copy layers, enabling multimodal control. Ham et al.~\shortcite{DBLP:journals/corr/abs-2302-12764} also integrate a modulation network into a diffusion model for a similar purpose. 
\yeA{
GLIGEN~\cite{DBLP:conf/cvpr/LiLWMYGLL23} introduces trainable gated transformer layers into the trained diffusion models to take in other modal inputs. 
}
SpaText~\cite{DBLP:journals/corr/abs-2211-14305} learns to generate images from spatio-textual representation. 
\yeA{
SceneComposer~\cite{DBLP:conf/cvpr/Zeng0ZLCK023} trains diffusion models that generate images from text feature pyramids. 
}
However, these methods require additional training or finetuning. 

In contrast, Directed diffusion~\cite{ma2023directed} is a training-free approach that directly adjusts the strength of cross-attention maps in the reverse diffusion process. However, this method can only handle bounding boxes as input. 
\yeA{Although the task is different, prompt-to-prompt~\cite{DBLP:journals/corr/abs-2208-01626} also focuses on controlling cross-attention maps for image editing.}
Paint-with-words in eDiff-i~\cite{DBLP:journals/corr/abs-2211-01324} can handle arbitrary-shaped masks by adding segmentation maps to cross-attention maps. However, adding uniform values to cross-attention maps often fails to generate mask-aligned images. 
MultiDiffusion~\cite{bar2023multidiffusion} also does not require additional training. This method performs a reverse diffusion process using different text input for each semantic region. In this method, bootstrapping encourages foreground objects to be generated inside semantic regions. However, generating each object independently during bootstrapping sometimes yields inconsistency with foregrounds and background. In addition, there is room for improvement in the accuracy of image generation according to input semantic regions. In Section \ref{sec:exp}, we demonstrate the effectiveness of our method through comparisons with these baselines.

\section{Method}
The goal of this study is to enable spatial control of text-to-image generation based on visual annotations. The user provides a text prompt $y$ and a semantic mask $S$ as input. Moreover, the user specifies correspondences $C$ between subsets of words in $y$ and semantic regions in $S$. The semantic mask $S$ is defined as $S=\{S_i\}^N_{i=1}$ ($N$ is the number of objects), where $S_i$ consists of pixel indices associated with an object $i$. We also define $C=\{C_i\}^N_{i=1}$, where $C_i$ consists of word indices associated with an object $i$.

\begin{figure}[t]
\centering
    \includegraphics[width=\linewidth]{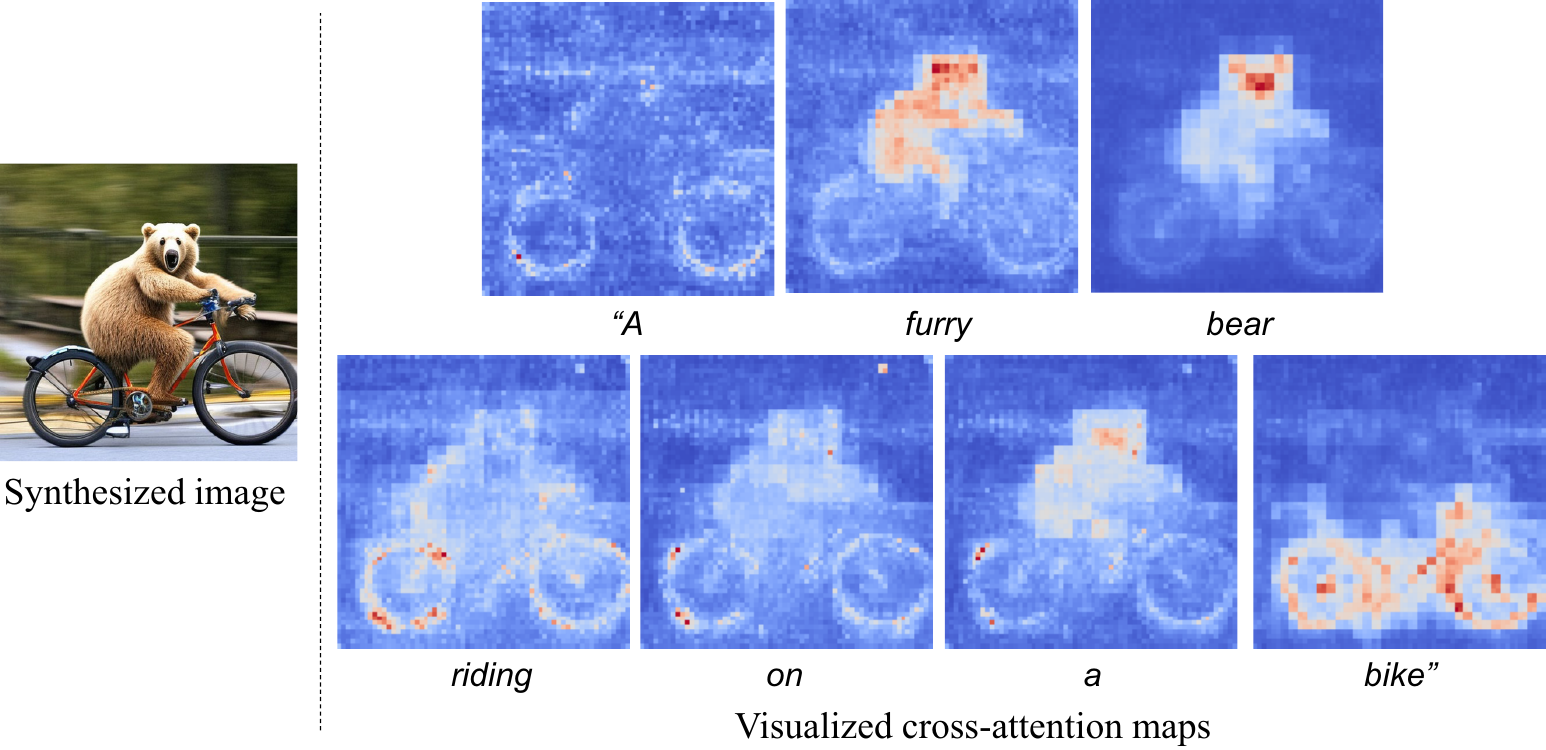}
\caption{Visualization of cross-attention maps estimated by Stable Diffusion. The redder pixels indicate stronger relationships with each word. The network attends to different pixels for each word. }
\label{fig:attention}
\end{figure}

For text-to-image generation, We adopt Stable Diffusion~\cite{Rombach_2022_CVPR}, a lightweight diffusion model that can produce high-quality and diverse images. Stable Diffusion generates images by iteratively denoising a noise map $z_t$ at each time step $t$ ($t=T, T-1,...,1$) using a text prompt $y$ as input. 
During this reverse diffusion process, we can obtain cross-attention maps from the intermediate cross-attention layers. As shown in Figure~\ref{fig:attention} and demonstrated in prompt-to-prompt~\cite{DBLP:journals/corr/abs-2208-01626}, the cross-attention maps indicate the relevance between given prompt words and pixels. The cross-attention map for each layer is computed by using the query $Q$ derived from a noise map $z_t$ and the key $K$ derived from a prompt $y$ as follows:
\begin{align}
    M = {\rm Softmax}\left(\frac{QK^T}{\sqrt{d}}\right),
\end{align}
where $d$ is a constant determined by the latent projection dimension of the key and query. 

Our main idea is to control cross attentions according to visual guidance. Therefore, our method is applicable to any model that incorporates a similar attention mechanism, not only Stable Diffusion. In Section \ref{sec:direct}, we first explore a very simple approach of directly assigning specific values to cross-attention maps. Next, in Section \ref{sec:guide}, we present masked-attention guidance, which indirectly adjusts corss-attention maps by modifying noise maps fed to the diffusion model, addressing the limitations of the first approach. 

\subsection{Masked-attention swapping}
\label{sec:direct}
Recently, Ma et al. proposed \yeA{Directed Diffusion~\shortcite{ma2023directed}} to spatially control the output of diffusion models according to user-provided bounding boxes. This method directly adjusts the strength of the cross-attention maps during the reverse diffusion process. Specifically, it adds values based on 2D Gaussian distribution to the cross-attentions for target words and inside bounding boxes. It also scales down the cross-attentions for the target words and outside the bounding boxes by a factor smaller than one. However, this method has difficulty in handling arbitrary-shaped masks. 

Instead, we explore a simple method called masked-attention swapping, which directly assigns specific values to the masked regions of cross-attention maps. Specifically, we swap the estimated cross-attention maps for target words with constant maps. Considering attention values at each pixel sum to one, we generate constant maps consisting of an inverse of $|C_i|$ inside target regions and zero outside those regions. This way, we evenly distribute constants over the target regions in the cross-attention maps for target words.

\begin{figure}[t]
\centering
    \includegraphics[width=\linewidth]{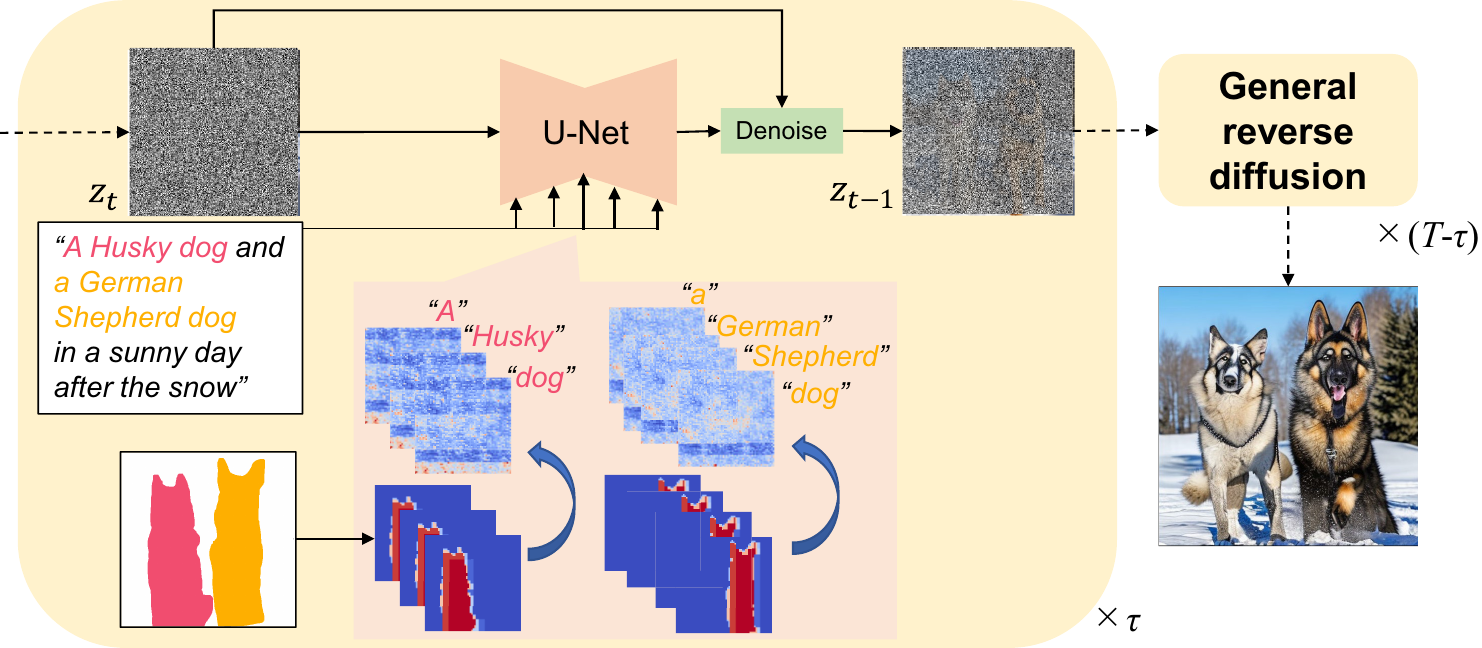}
\caption{Overview of masked-attention swapping. At each reverse diffusion step, we replace estimated attention maps with constant maps computed based on given semantic regions. This procedure is iterated $\tau$ steps, and the general reverse diffusion process is performed remaining $T-\tau$ steps ($T$ is the total number of the reverse diffusion steps). Note that we omit the encoder and decoder of Stable Diffusion for compression from the figure to save space. }
\label{fig:method1}
\end{figure}

Figure~\ref{fig:method1} illustrates the pipeline of masked-attention swapping. The U-Net takes as input a noisy image $z_t$  at each time step $t$ (t=T, T-1, …, 1) and estimates its noise. It then uses the estimated noise to denoise $z_t$ and feeds it back to the U-Net. In this iterative process of reverse diffusion, we swap the intermediate cross-attention maps with constant maps computed from an input semantic mask $S$. Following \yeA{Directed Diffusion}~\shortcite{ma2023directed}, we only apply attention swapping for the first $\tau$ iterations of the reverse diffusion process. This is because later iterations affect not scene layout but fine details. 

\begin{figure}[t]
\centering
    \includegraphics[width=\linewidth]{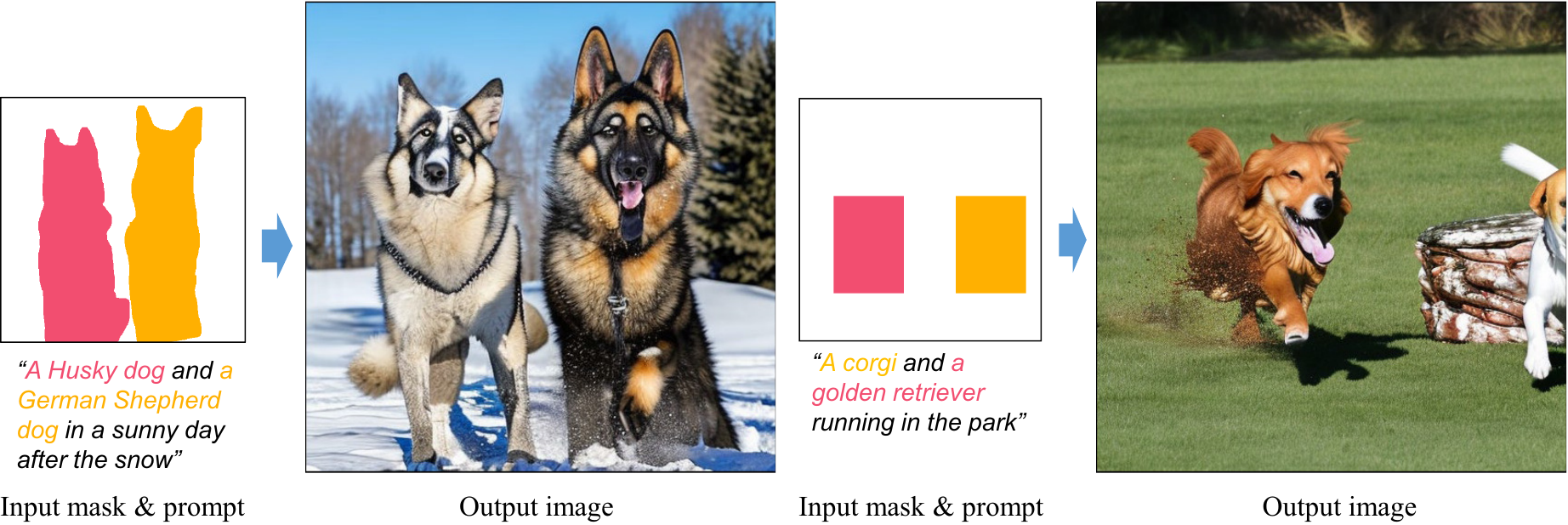}
\caption{Results of masked-attention swapping. As shown in the right example, this approach sometimes fails to generate objects in the specified regions. }
\label{fig:direct_failure}
\end{figure}

As shown in Figure~\ref{fig:direct_failure}, masked-attention swapping enables some spatial control (left). However, we also observe the case where it yields large misalignment with the semantic regions (right). We identify two possible causes for this issue. First, our method injects spatially uniform values into the target regions of cross-attention maps. In contrast, as shown in Figure~\ref{fig:attention}, the actual cross-attention maps are spatially varying. Although \yeA{Directed Diffusion}~\shortcite{ma2023directed} employ 2D Gaussian distribution to assign spatially varying values, this approach has a limited ability to capture attention distribution. Second, masked-attention swapping also injects uniform values into attentions across all target words, but each word should relate to a different region. However, manually specifying attention values for each pixel and word is impractical. 

\subsection{Masked-attention guidance}
\label{sec:guide}
To address the issues mentioned in Section~\ref{sec:direct}, we propose masked-attention guidance. Inspired by pix2pix-zero~\cite{DBLP:journals/corr/abs-2302-03027}, instead of directly manipulating cross-attention maps, we modify noise maps fed to the diffusion model to control the cross-attention maps. Masked-attention guidance does not specify specific attention values but only controls increases and decreases in attention. The advantage of this approach is that it can determine which words and pixels to be attended to by leveraging the capability of attention estimation in well-trained diffusion models. 

\begin{figure}[t]
\centering
    \includegraphics[width=\linewidth]{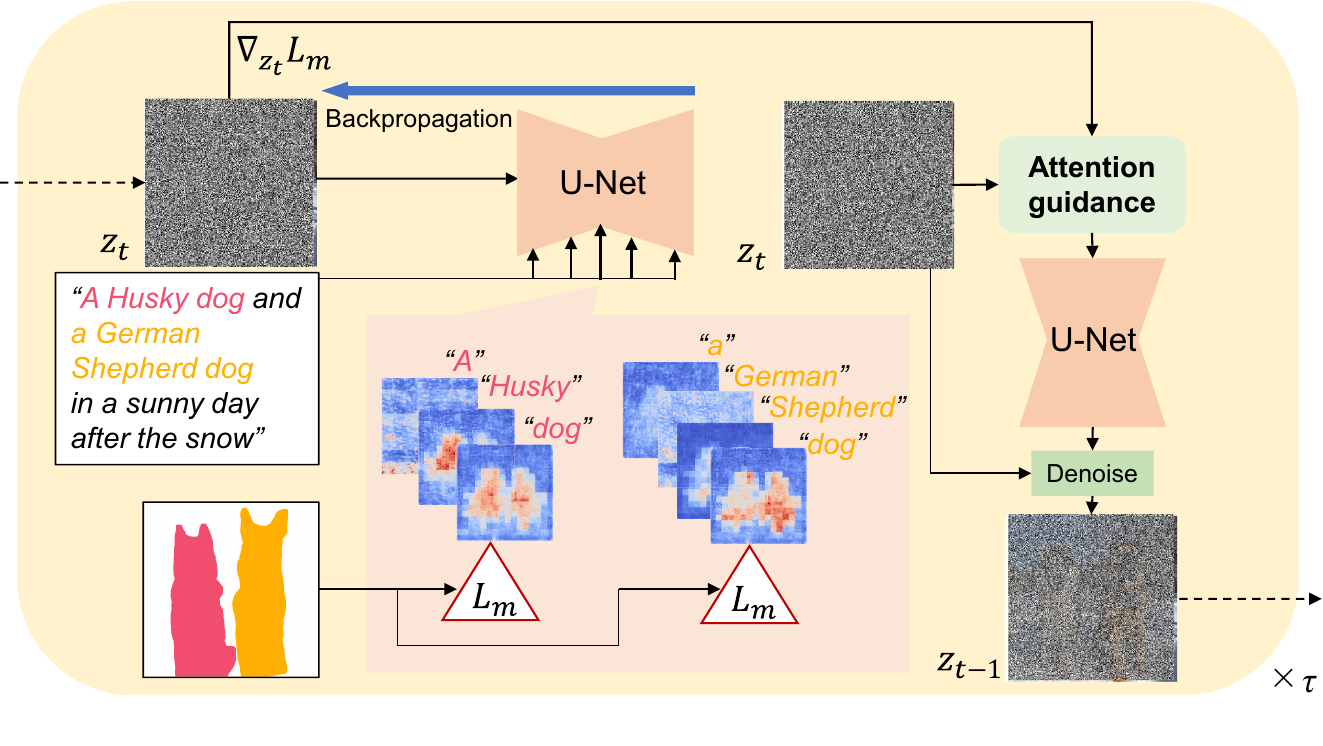}
\caption{
Overview of masked-attention guidance. At each reverse diffusion step, we first compute cross-attention maps from a noise map $z_t$ and compute the masked-attention loss $\mathcal{L}_m$. We then perform backpropagation to compute $\nabla_{z_t} \mathcal{L}_m$. Using $\nabla_{z_t} \mathcal{L}_m$, we apply masked-attention guidance to $z_t$ and feed the guided noise map to the U-Net to obtain $z_{t-1}$. The procedure is iterated $\tau$ steps, and the general reverse diffusion process (omitted in the figure to save space) is performed remaining $T-\tau$ steps ($T$ is the total number of the reverse diffusion steps). Note that we omit the encoder and decoder of Stable Diffusion for compression from the figure to save space.
}
\label{fig:method2}
\end{figure}

\begin{figure*}[t]
\centering
    \includegraphics[width=\linewidth]{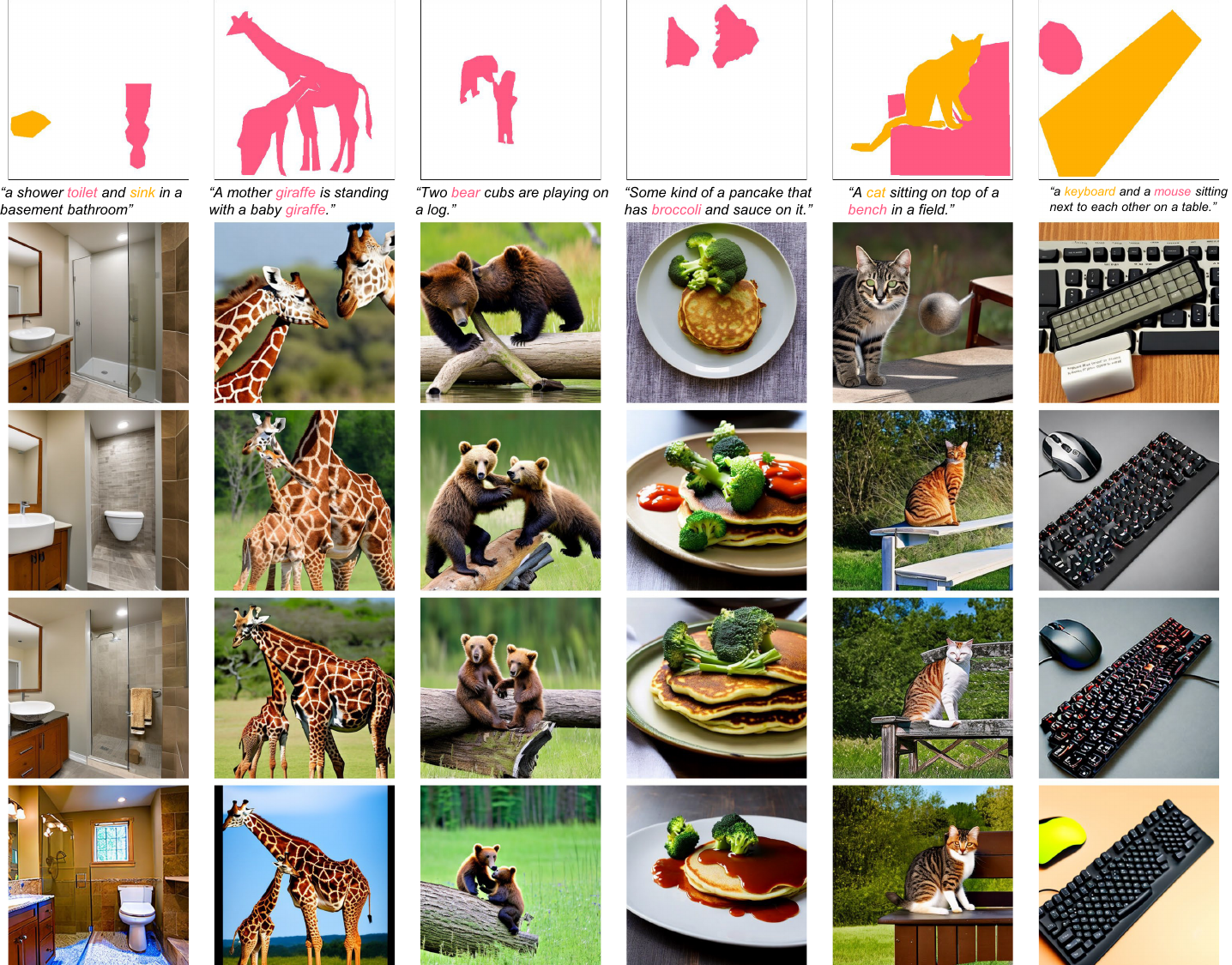}
\caption{Qualitative comparison on the COCO dataset~\cite{DBLP:conf/eccv/LinMBHPRDZ14}. From top to bottom: input masks and prompts, results of MultiDiffusion~\cite{bar2023multidiffusion}, paint-with-words~\cite{DBLP:journals/corr/abs-2211-01324}, masked-attention swapping (Section~\ref{sec:direct}), and our method (masked-attention guidance in Section~\ref{sec:guide}). }
\label{fig:coco1}
\end{figure*}

Pix2pix-zero~\cite{DBLP:journals/corr/abs-2302-03027} tackles a problem of editing real images using text input while preserving input image layouts. This method manipulates noise maps with L2 loss-based guidance that encourages cross-attention maps before and after editing to be close. However, in our problem setting, we cannot use this approach because we have no cross-attention maps computed from real images before editing. Our masked-attention guidance, therefore, controls attentions without using such ground-truth cross-attention maps. Specifically, given a semantic mask $S$ and correspondences $C$ between semantic regions and prompt words, the masked-attention loss $\mathcal{L}_m$ is defined as:
\begin{align}
\mathcal{L}_m = -\sum_{i=1}^{N}\sum_{w\in C_i}\sum_{p\in S_i}M_{p,w}+\lambda \sum_{i=1}^{N}\sum_{w\in C_i}\sum_{p\in \bar{S}_i}M_{p,w}, 
\label{eq:mal}
\end{align}
where $M_{p,w}$ is a value of a cross-attention map $M$ for word $w$ at pixel $p$, and $\lambda$ is a balancing weight. This loss has a role to control the increase and decrease of attentions for specific regions and words rather than getting them closer to specific values. Namely, the first term increases the attentions for words and pixels of an object $i$, and the second term decreases the attentions for words and outside pixels of an object $i$. Then, masked-attention guidance updates $z_t$ as follows:  
\begin{align}
z_t \leftarrow z_t - \alpha \nabla_{z_t} \mathcal{L}_m,
\label{eq:mag}
\end{align}
where $\alpha$ is a guidance scale, and $\nabla_{z_t} \mathcal{L}_m$ is computed by backpropagating the loss $\mathcal{L}_m$ with respect to a noise image $z_t$. Again, masked-attention guidance only encourages the attention to increase or decrease for specific pixels and words. Therefore, how much the network attends to which pixels and words is determined by leveraging well-trained diffusion models. We validate the effectiveness of this approach by analyzing estimated cross-attention maps in Section~\ref{sec:eval_qualitative}. 

Figure~\ref{fig:method2} shows the pipeline of masked-attention guidance. In a reverse diffusion step, we first feed a noise image $z_t$ to the U-Net and obtain cross-attention maps. Next, we calculate the masked-attention loss $\mathcal{L}_m$ using Equation (\ref{eq:mal}) for the obtained cross-attention maps. We then compute $\nabla_{z_t} \mathcal{L}_m$ via backpropagation and update the noise image $z_t$ using Equation (\ref{eq:mag}). We feed the updated $z_t$ to the U-Net again and denoise $z_t$. At this time, other guidance methods (e.g., classifier-free guidance~\cite{DBLP:journals/corr/abs-2207-12598} can also be applied. We iterate this procedure for $\tau$ steps and perform the general reverse diffusion process for the remaining $T-\tau$ steps. 

\begin{figure}[t]
\centering
    \includegraphics[width=\linewidth]{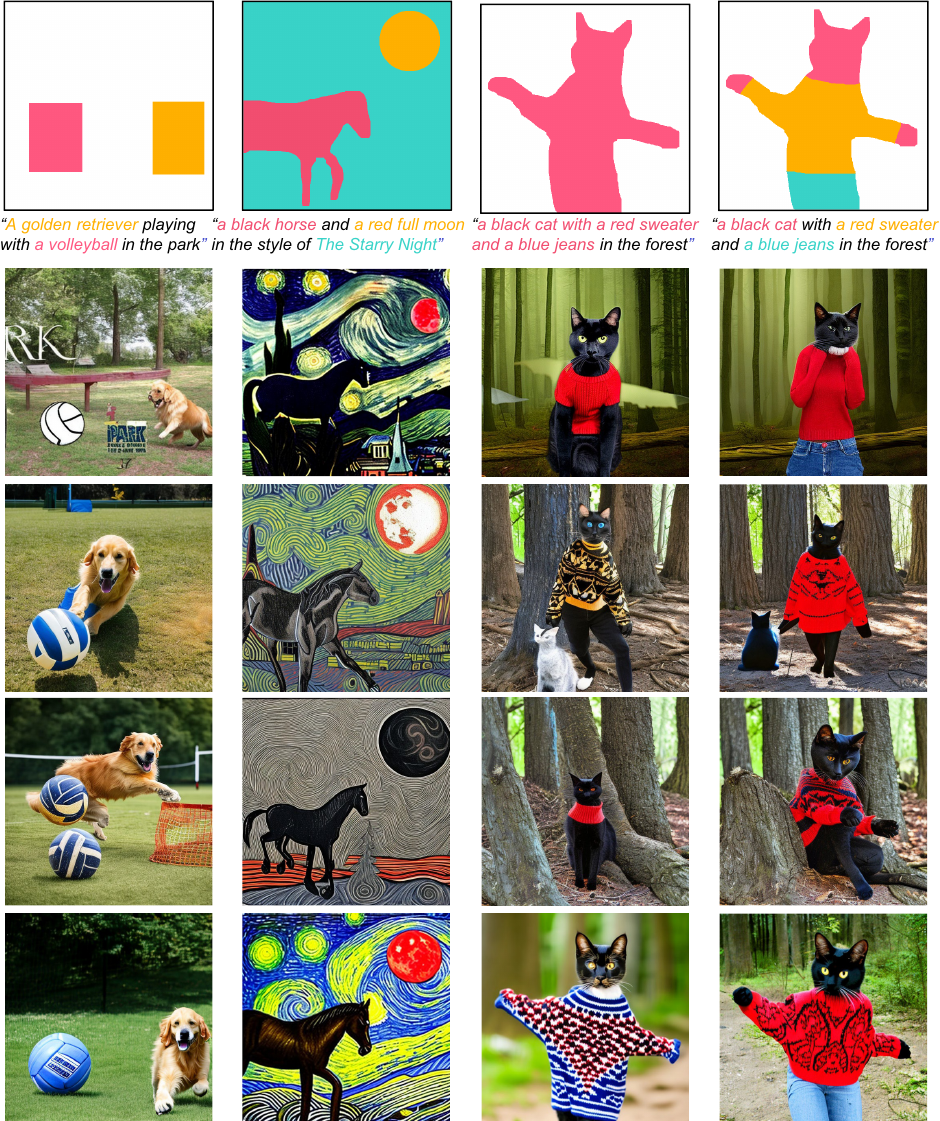}
\caption{Qualitative comparison. From top to bottom: input masks and prompts, results of MultiDiffusion~\cite{bar2023multidiffusion}, paint-with-words~\cite{DBLP:journals/corr/abs-2211-01324}, masked-attention swapping (Section~\ref{sec:direct}), and our method (masked-attention guidance in Section~\ref{sec:guide}). }
\label{fig:res}
\end{figure}

\section{Experiments}
\label{sec:exp}
\paragraph{Implementation details}
We implemented our method using Python and PyTorch and ran our program on an NVIDIA RTX A5000. We used Stable Diffusion~v2.0~\cite{Rombach_2022_CVPR} for the diffusion model and the DDIM sampler~\cite{DBLP:conf/iclr/SongME21} with 50 sampling steps for the reverse diffusion process. 
We applied masked-attention swapping or masked-attention guidance to all cross-attention layers of the diffusion model. To do so, we resized input masks to the same size as the cross-attention maps in each layer. 
We performed these approaches in the first $30\%$ of the total denoising steps. 
The masked-attention guidance scale $\alpha$ and its loss weight $\lambda$ were set to 0.08 and 0.5, respectively, in all experiments except for the ablation study in Section~\ref{sec:ablation}. Our method with masked-attention guidance took about 9 seconds, whereas Stable Diffusion without our guidance took about 5 seconds. 

\subsection{Quantitative evaluation}
\label{sec:eval_quantitative}
\paragraph{Dataset.} 
Following MultiDiffuion~\cite{bar2023multidiffusion}, we conducted quantitative evaluation using the COCO dataset~\cite{DBLP:conf/eccv/LinMBHPRDZ14}. This dataset contains images with global captions explaining the entire images and instance masks for 80 classes. We first filtered images containing 2-4 instances except for a person from 5,000 images in the 2017 validation set. This procedure is similar to MultiDiffuion. Because our method needs to take the correspondence between each instance mask and the words in the captions, we further filtered images with captions that contain the classes of corresponding instance masks. This procedure resulted in 348 sets of captions and masks. We performed text-to-image generation from these inputs using our method and baselines. Finally, we applied the off-the-shelf instance segmentation method~\cite{DBLP:conf/cvpr/ChengMSKG22} to the generated images and computed mean Intersection over Union (mIoU) between the estimated segmentation results and the input instance masks. In addition, we used $\rm{FID_{CLIP}}$~\cite{Kynkaanniemi2022} to evaluate the image quality via comparison between the COCO ground-truth images and generated images. 

\paragraph{Comparison with baselines.}
We compared our method with MultiDiffusion~\cite{bar2023multidiffusion} and paint-with-words in eDiff-i~\cite{DBLP:journals/corr/abs-2211-01324}, which are recent training-free approaches for spatially controlling text-to-image diffusion models. For MultiDiffusion, we applied bootstrapping to $20\%$ (the same percentage as their paper) of all denoising steps. For paint-with-words, we implemented it in Stable Diffusion~\cite{Rombach_2022_CVPR}. In this method, we need to determine the weight parameter $w'$, which controls the value added to cross-attention maps. We searched for the value resulting in the best mIoU from 0 to 1 in increments of 0.1 and set $w'$ as 0.3. 

Table ~\ref{tab:quant2} shows the quantitative comparison. The results show that the existing methods (i.e., MultiDiffusion and paint-with-words) is worse than two of our methods in mIoU. Our masked-attention swapping shows the better scores for each metric, but improvement of mIoU is modest. Meanwhile, our masked-attention guidance significantly outperforms these baselines in mIoU \yeA{and} can generate images more faithful to input masks. \yeA{Although our method shows the slightly worse $\rm{FID_{CLIP}}$ score, we can obtain visually plausible results, as demonstrated in Section~\ref{sec:eval_qualitative}.} Note that image quality and mask alignment is a trade-off in our guidance. We used the guidance scale $\alpha=0.08$ in this experiment, but adjusting $\alpha$ can also improve the image quality with modestly decreasing mIoU, as discussed in the ablation study in Section~\ref{sec:ablation}. 

\begin{table}[]
\caption{Quantitative comparison between our method and the baselines.}
\label{tab:quant2}
\center
\begin{tabular}{l|cc}
Method & mIoU $\uparrow$             & $\rm{FID_{CLIP}}$ $\downarrow$  \\ \hline
MultiDiffusion~\cite{bar2023multidiffusion}                       & 0.255 $\pm$ 0.225 & 24.0 \\
Paint-with-words~\cite{DBLP:journals/corr/abs-2211-01324}                      & 0.317 $\pm$ 0.273 & 22.2 \\
Masked-attention swapping           & 0.330 $\pm$ 0.297 & 21.8 \\
Ours (masked-attention guidance)     & 0.457 $\pm$ 0.300 & 24.9
\end{tabular}
\end{table}

\begin{table}[]
\caption{\yeA{Quantitative comparison between our masked-attention guidance using bounding boxes or semantic masks as input.}}
\label{tab:diff_inputs}
\center
\yeA{
\begin{tabular}{l|cc}
Method & mIoU $\uparrow$             & $\rm{FID_{CLIP}}$ $\downarrow$  \\ \hline
Ours w/ bounding boxes          & 0.318 $\pm$ 0.251 & 29.1 \\
Ours w/ semantic masks     & 0.457 $\pm$ 0.300 & 24.9
\end{tabular}
}
\end{table}

\paragraph{\yeA{Effectiveness of semantic masks. }}
\yeA{
We validated the effectiveness of using semantic masks as input for our masked-attention guidance. To do so, we created bounding boxes from the masks in the COCO dataset. Using these bounding boxes as input, we generated images and performed an evaluation for ground-truth images and segmentation maps in the COCO dataset. As shown in Table~\ref{tab:diff_inputs}, the results from semantic masks show better scores, which indicate that semantic masks allow more accurate control than bounding boxes. 
}

\begin{figure}[t]
\centering
    \includegraphics[width=\linewidth]{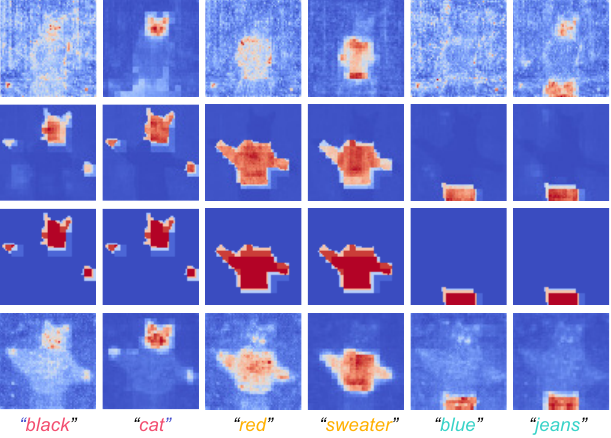}
\caption{Analyzing cross-attention maps obtained by \yeA{each} method. Cross-attention maps are visualized after performing $30\%$ of the total denoising steps. \yeA{From top to bottom: cross-attention maps of MultiDiffusion~\cite{bar2023multidiffusion}, paint-with-words~\cite{DBLP:journals/corr/abs-2211-01324}, masked-attention swapping (Section~\ref{sec:direct}), and our method (masked-attention guidance in Section~\ref{sec:guide}).}}
\label{fig:att_analysis}
\end{figure}

\subsection{Qualitative evaluation}
\label{sec:eval_qualitative}
\begin{figure*}[t]
\centering
    \includegraphics[width=\linewidth]{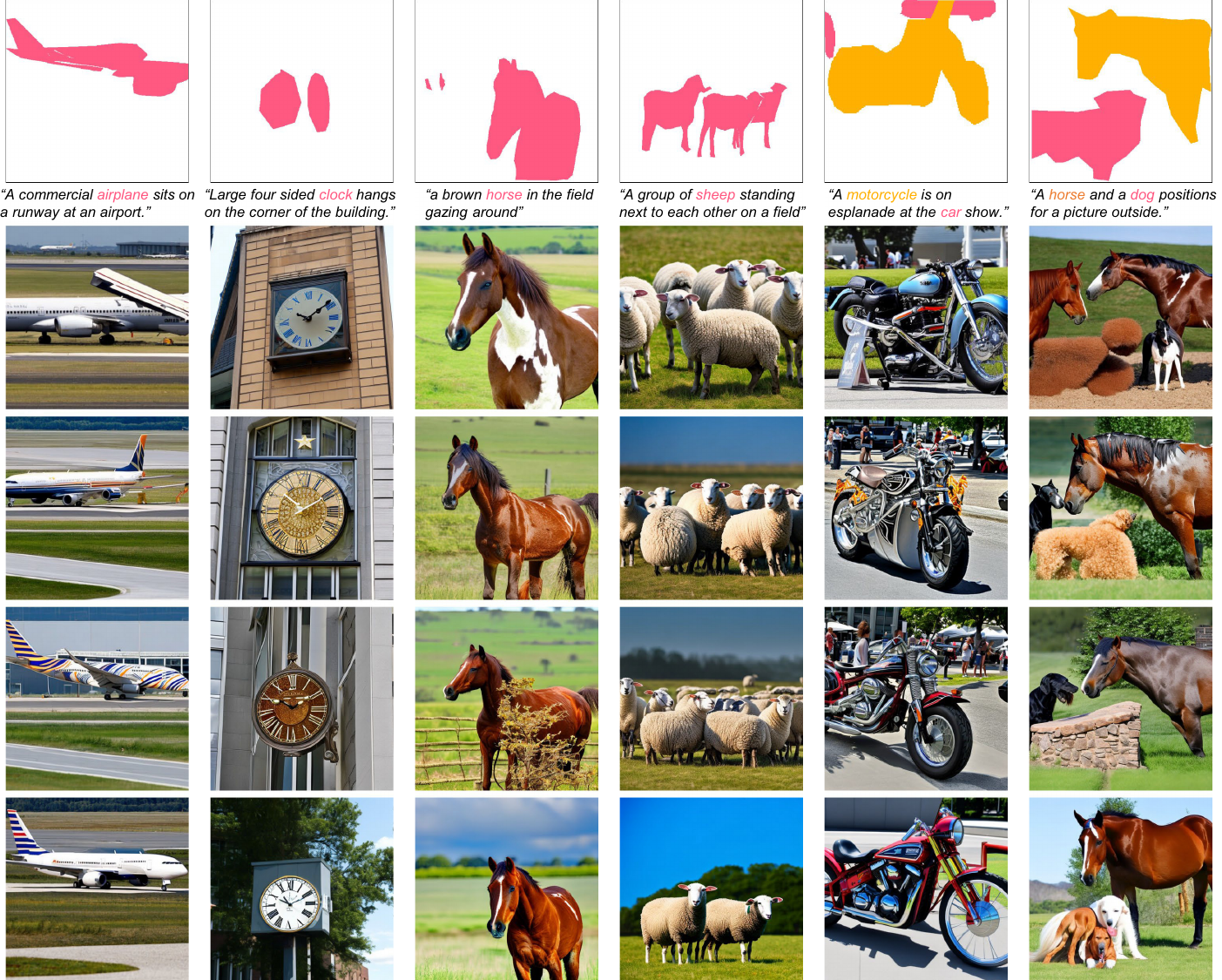}
\caption{
Additional qualitative comparison on the COCO dataset~\cite{DBLP:conf/eccv/LinMBHPRDZ14}. From top to bottom: input masks and prompts, results of MultiDiffusion~\cite{bar2023multidiffusion}, paint-with-words~\cite{DBLP:journals/corr/abs-2211-01324}, masked-attention swapping (Section~\ref{sec:direct}), and our method (masked-attention guidance in Section~\ref{sec:guide}). 
}
\label{fig:coco2}
\end{figure*}

\begin{figure*}[t]
\centering
    \includegraphics[width=\linewidth]{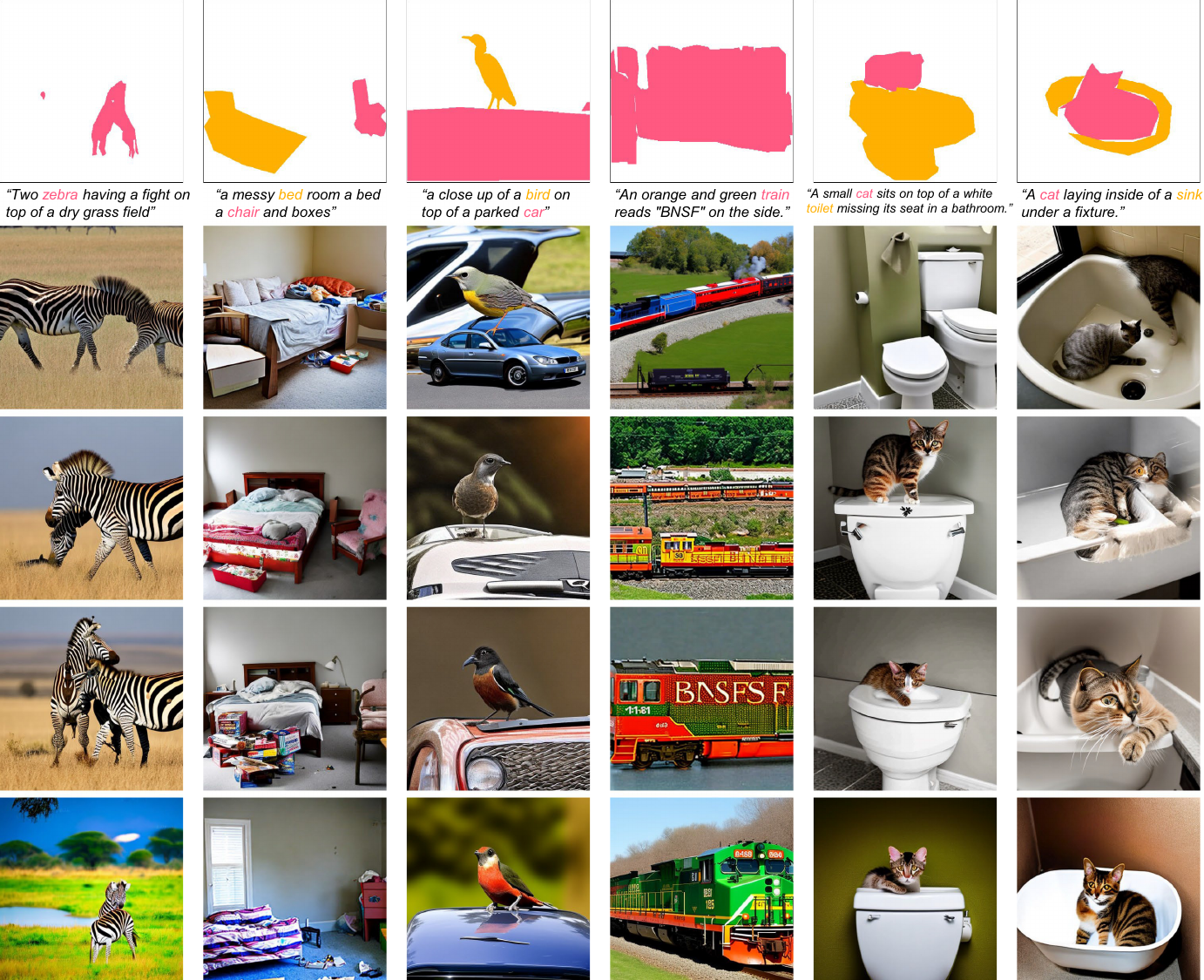}
\caption{
Additional qualitative comparison on the COCO dataset~\cite{DBLP:conf/eccv/LinMBHPRDZ14}. From top to bottom: input masks and prompts, results of MultiDiffusion~\cite{bar2023multidiffusion}, paint-with-words~\cite{DBLP:journals/corr/abs-2211-01324}, masked-attention swapping (Section~\ref{sec:direct}), and our method (masked-attention guidance in Section~\ref{sec:guide}).
}
\label{fig:coco3}
\end{figure*}

\paragraph{Comparison with baselines}
Figure~\ref{fig:coco1} shows a qualitative comparison between our method and the baseline methods on the COCO dataset.
In the results of our method, the generated images are overall better aligned with the input masks compared to the baseline methods.
For example, the results of the first column show that some of the baselines do not generate a toilet, whereas our method does.
In the second and third columns, our method generates images more aligned with the input masks.
In the fifth and sixth columns, masked-attention swapping also obtains the images faithful to the masks, yet with noticeable artifacts.
In contrast, our method obtains more natural results.  
These qualitative results are also consistent with the quantitative results. Figures~\ref{fig:coco2} and~\ref{fig:coco3} also provide more results.

\begin{figure}[t]
\centering
    \includegraphics[width=\linewidth]{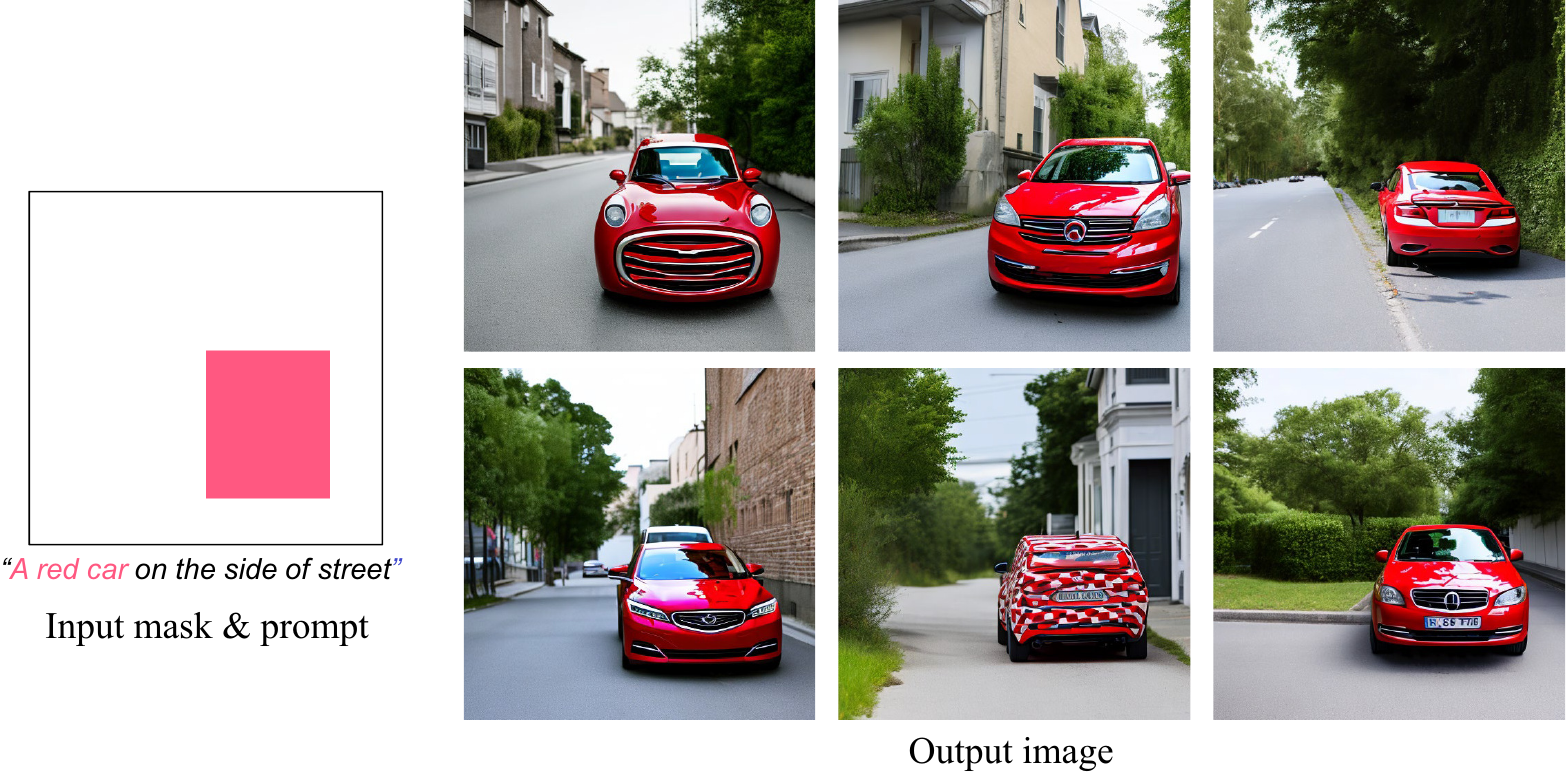}
\caption{Diverse results generated by our method with a single semantic mask and prompt. }
\label{fig:diversity}
\end{figure}

Furthermore, Figure~\ref{fig:res} compares the results generated using several challenging inputs not in the COCO dataset. 
In the first column, MultiDiffusion also produces an image according to the bounding boxes, but it creates inconsistent foregrounds and background (i.e., a cartoon volleyball exists in a real photograph). 
The second column shows that our method generates an image aligned with the mask while also reflecting the style of "\textit{The Starry Night}" in the background. 
The third column is a challenging case with many words for a single semantic region.  
In this case, our method generates an image more faithful to the mask than the baselines but struggles to estimate appropriate garment colors in the specified region. Interestingly, as shown in the fourth column, this problem can be alleviated by specifying additional semantic regions.

\begin{figure}[t]
\centering
    \includegraphics[width=\linewidth]{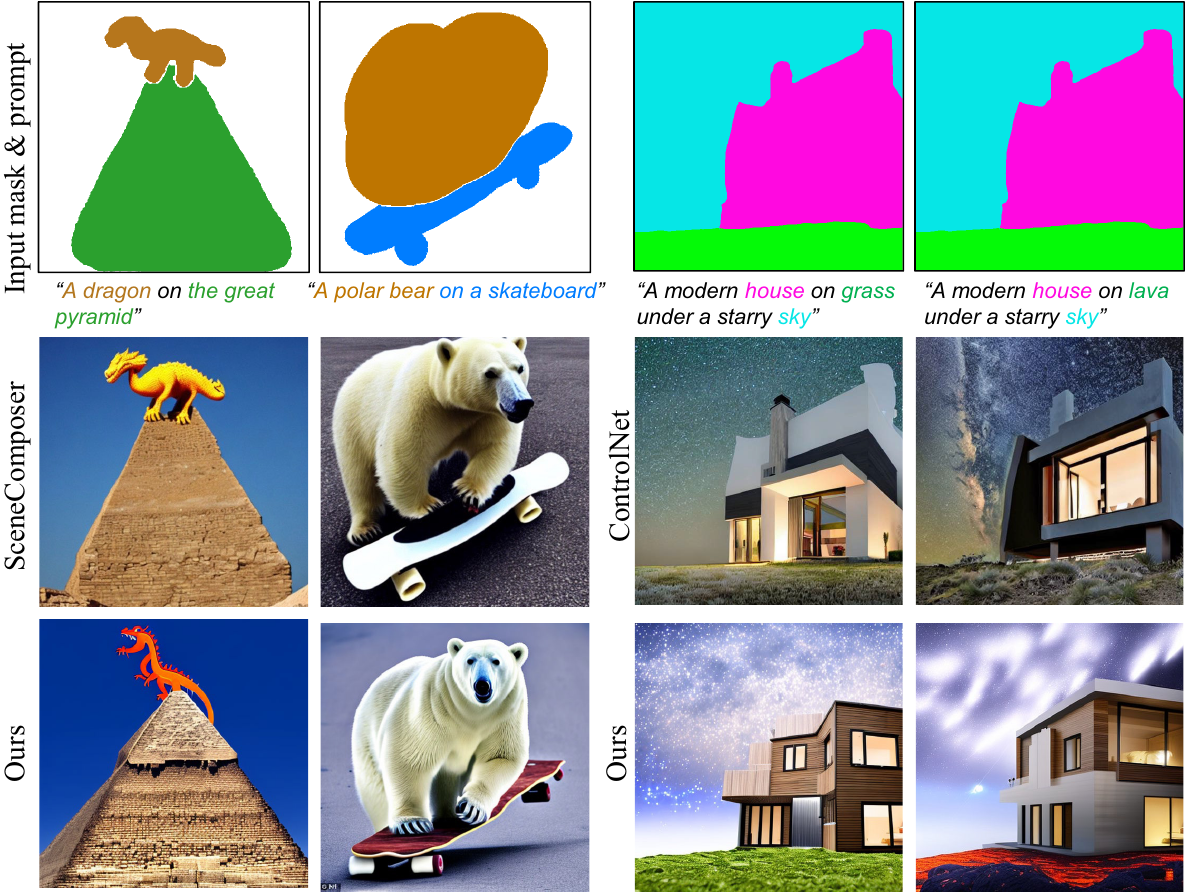}
\caption{\yeA{Qualitative comparison with the supervised methods, SceneComposer~\cite{DBLP:conf/cvpr/Zeng0ZLCK023} and ControlNet~\cite{DBLP:journals/corr/abs-2302-05543}.} }
\label{fig:eval_supervised}
\end{figure}

\paragraph{\yeA{Comparison with supervised methods. }}
\yeA{
Although the supervised approaches, such as ControlNet~\cite{DBLP:journals/corr/abs-2302-05543} and SceneComposer~\cite{DBLP:conf/cvpr/Zeng0ZLCK023}, require further training or finetuning on curated datasets, we compared with these existing methods to validate the potential of our training-free method. As shown in Figure~\ref{fig:eval_supervised}, despite adopting the training-free approach, our method produces good results, although not as good as the supervised methods. Meanwhile, the supervised methods can precisely generate images even for small regions (e.g., dragon and chimney). However, SceneComposer needs to train diffusion models from scratch. For ControlNet, class labels assigned to semantic regions are limited to predefined ones in training datasets (e.g., 150 classes in ADE20K~\cite{DBLP:conf/cvpr/ZhouZPFB017}). For example, in Figure~\ref{fig:eval_supervised}, the green region in the semantic mask has a "\textit{grass}" class, but "\textit{lava}" is not contained in the predefined classes. As a result, ControlNet cannot reflect "\textit{lava}" in the green region, whereas our method can assign arbitrary texts to semantic regions. 
}

\paragraph{Analyzing cross-attention maps.}
Figure~\ref{fig:att_analysis} shows the visualization of cross-attention maps estimated via the reverse diffusion process with \yeA{each method}. \yeA{These cross-attention maps are obtained from the input in the fourth column in Figure~\ref{fig:res}}. 
\yeA{
In these results, MultiDiffusion cannot sufficiently attend to the regions related to each word. Although paint-with-words and masked-attention swapping do not suffer from this problem, they uniformly attend to each pixel and word. In contrast to these methods, our masked-attention guidance
}
attends to different regions with various intensities for each word. This suggests that masked-attention guidance can effectively guide attention estimation by leveraging the inference ability of the well-trained diffusion model. 

\paragraph{Diversity.}
Diffusion models can generate various images from different initial noise maps. 
As shown in Figure~\ref{fig:diversity}, our method can also generate diverse images according to input texts and semantic masks. 

\subsection{Ablation study}
\label{sec:ablation}
\begin{figure}[t]
\centering
    \includegraphics[width=\linewidth]{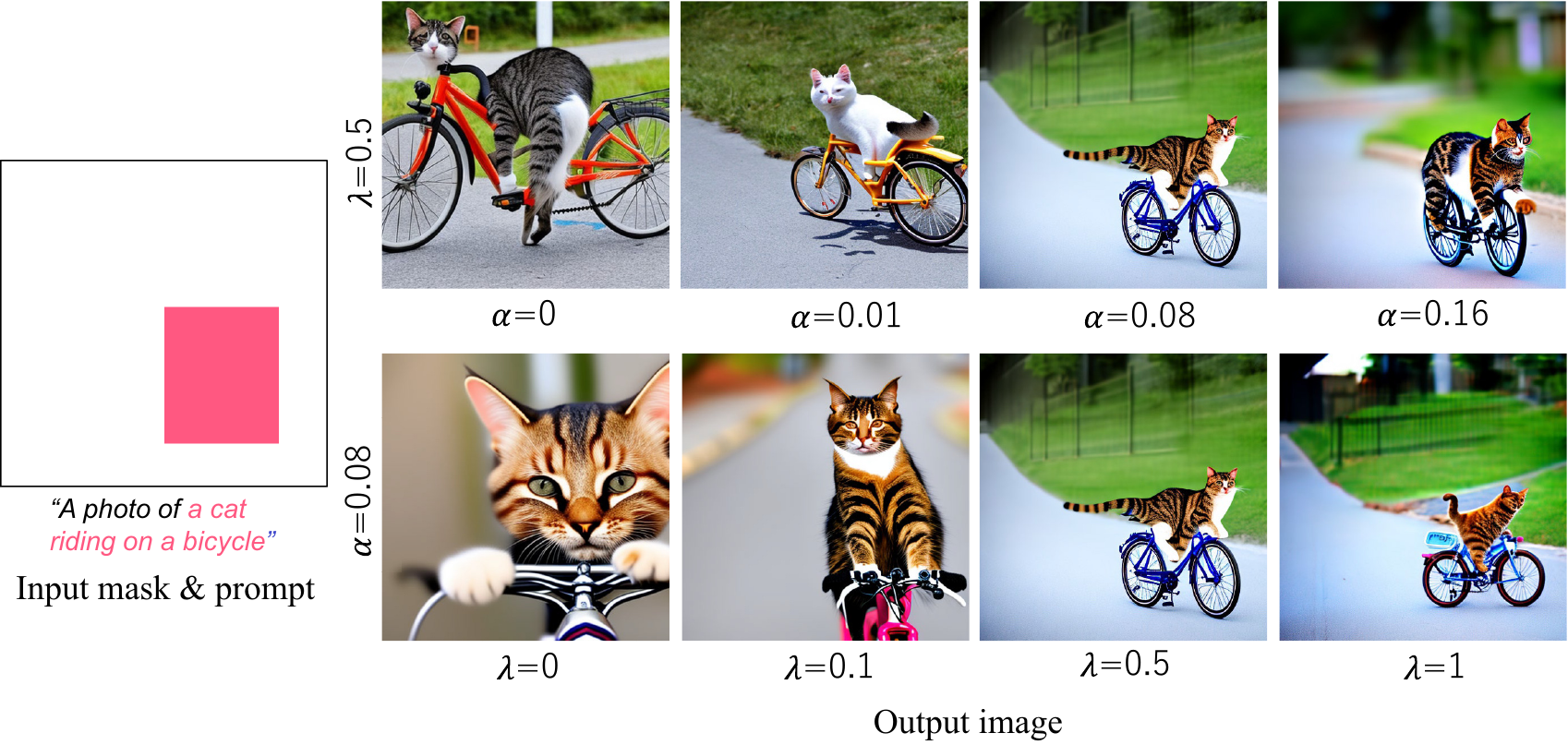}
\caption{Results generated with different masked-attention guidance scale~$\alpha$ and its loss weight $\lambda$. }
\label{fig:ablation}
\end{figure}

We conducted an ablation study to investigate the effectiveness of our masked-attention guidance.
Figure ~\ref{fig:ablation} shows how different values of the masked-attention guidance scale $\alpha$ and its loss weight $\lambda$ affect results generated by our method. 
When $\alpha=0$, the semantic mask is not considered completely because masked-attention guidance is not applied. 
The larger $\alpha$, the more accurately our method generates the cats that fit inside the bounding box.
On the other hand, $\lambda$ determines the strength of the loss for avoiding generating objects outside specified regions.
If $\lambda$ is too small, our method generates the cats that largely protrude outside the target area. 
In addition, Table~\ref{tab:quant3} shows quantitative comparison of our masked attention guidance with different guidance scale $\alpha$. We can see that image quality and mask alignment is basically a trade-off depending on $\alpha$. This relationship is similar to one between FID and Inception score in classifier guidance~\cite{DBLP:conf/nips/DhariwalN21} and classifier-free guidance~\cite{DBLP:journals/corr/abs-2207-12598}. 

\begin{figure*}[t]
\centering
    \includegraphics[width=\linewidth]{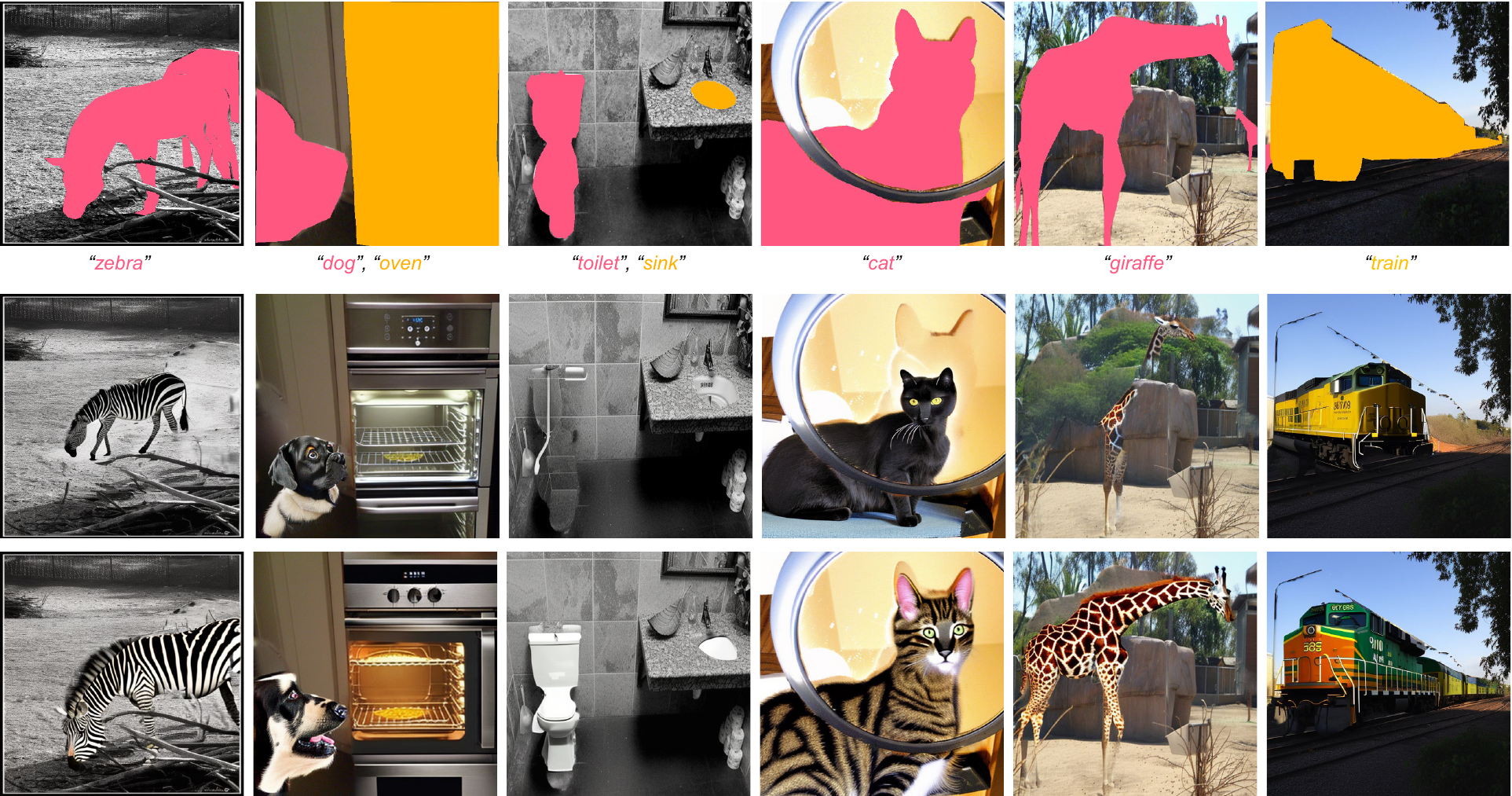}
\caption{
Qualitative comparison between Blended Latent Diffusion~\cite{DBLP:journals/corr/abs-2206-02779} with and without our masked-attention guidance. 
From top to bottom: input masks and prompts, results of Blended Latent Diffusion, and Blended Latent Diffusion with our masked-attention guidance. }
\label{fig:inpaint}
\end{figure*}

\subsection{Application}
\label{sec:application}
Our masked-attention guidance is not only applicable to the task of text-to-image for generating images from scratch but also text-guided image editing for manipulating real images with textual input. 
Blended Latent Diffusion~\cite{DBLP:journals/corr/abs-2206-02779} is a state-of-the-art method in text-guided image editing.
This method generates objects in mask regions according to textual input by performing denoising only on the mask while preserving the noise outside the mask during the diffusion process. We can incorporate our masked-attention guidance into these denoising steps. We empirically set the guidance scale $\alpha$ as $0.01$. Quantitative and qualitative comparisons in Table~\ref{tab:quant4} and Figure~\ref{fig:inpaint} demonstrate that our method improves the performance of Blended Latent Diffusion. 

\begin{table}[]
\caption{Quantitative comparison with different masked-attention guidance scale $\alpha$. }
\label{tab:quant3}
\center
\begin{tabular}{c|cc}
 Guidance scale $\alpha$ & mIoU $\uparrow$                  & $\rm{FID_{CLIP}}$ $\downarrow$ \\ \hline
 $0.01$                                 & 0.329 $\pm$ 0.256 & 23.0 \\
 $0.04$                                 & 0.427 $\pm$ 0.286 & 23.6 \\
 $0.08$                                 & 0.457 $\pm$ 0.300 & 24.9 \\
 $0.16$                                 & 0.442 $\pm$ 0.315 & 27.6
\end{tabular}
\end{table}

\begin{table}[]
\caption{
Quantitative comparison between Blended Latent Diffusion~\cite{DBLP:journals/corr/abs-2206-02779} with and without our masked-attention guidance. 
}
\label{tab:quant4}
\center
\begin{tabular}{l|cc}
 Method                          & mIoU $\uparrow$              & $\rm{FID_{CLIP}}$ $\downarrow$    \\ \hline
 Blended Latent Diffusion~\cite{DBLP:journals/corr/abs-2206-02779}        & 0.406 $\pm$ 0.291 & 15.5 \\
 Blended Latent Diffusion~\cite{DBLP:journals/corr/abs-2206-02779} w/ours & 0.519 $\pm$ 0.303 & 14.8
\end{tabular}
\end{table}

\section{Conclusions}
We tackled a problem of spatially controlling diffusion models for text-to-image synthesis without additional training. To control cross-attention maps that capture the correspondence between pixels and prompt words, we first tried masked-attention swapping. This approach swaps cross-attention maps with constant maps generated according to given semantic regions at each reverse diffusion step. However, because the swapped maps are far from actual cross-attention maps, the generated images are not aligned well with the target regions. 
\yeA{Some existing training-free methods also suffer from a similar problem. 
To overcome this problem,}
we proposed masked-attention guidance, which manipulates the noise map so that the attentions to target words and inside (outside) target regions become larger (smaller). This guidance itself only encourages cross-attention maps to increase or decrease and leverages the ability of well-trained diffusion models to decide how much the network attends to which words and pixels. Quantitative evaluation on the COCO dataset \yeA{and various qualitative results} shows that our method can generate images more faithful to semantic masks than baselines. \yeA{In addition, we applied our method to the task of text-guided image editing.}

\paragraph{Limitations and future work.}
Our method has some limitations. As can be seen from some qualitative results, our method does not generate images that are completely aligned with given semantic masks. For example, we struggle to handle small regions like cat tails. 
One possible reason is that Stable Diffusion processes input images in a low-dimensional latent space for computational efficiency. Integrating our method into other diffusion models may allow for more precise control. 
\yeA{Another limitation is that our method cannot explicitly discriminate between multiple instance masks corresponding to identical words. As a result, our method cannot guarantee that the same number of objects as instance masks are generated. 
}
\yeA{Finally,} our method has a common problem with other guidance-based approaches, such as classifier guidance. Namely, it sometimes yields unnatural images when using a too-strong guidance scale $\alpha$. 
Nevertheless, our method performs better than baselines, and we expect various extensions of our approach in the future. For example, extending our method to handle other visual guidance like sparse scribbles is an exciting direction. 

\bibliographystyle{ACM-Reference-Format}
\bibliography{main}

\end{document}